%% file: main.tex
\documentclass[runningheads]{llncs}

\usepackage{eccv}

\usepackage{eccvabbrv}

\usepackage{graphicx}
\usepackage{float}
\usepackage{booktabs}
\usepackage{placeins}

\usepackage{enumitem}
\usepackage{makecell}
\usepackage{algorithm}
\usepackage{algorithmic}

\usepackage[accsupp]{axessibility}  % Improves PDF readability for those with disabilities.

\usepackage{hyperref}

\usepackage{orcidlink}

\begin{document}

\title{PiCo: Active Manifold Canonicalization for Robust Robotic Visual Anomaly Detection} 

\titlerunning{Abbreviated paper title}

% Second Author\inst{2,3}\orcidlink{1111-2222-3333-4444} \and
% Third Author\inst{3}\orcidlink{2222--3333-4444-5555}}

% % TODO FINAL: Replace with an abbreviated list of authors.

% % TODO FINAL: Replace with your institution list.
% Springer Heidelberg, Tiergartenstr.~17, 69121 Heidelberg, Germany

\author{
Teng Yan\inst{1} \and
Binkai Liu\inst{1} \and
Shuai Liu\inst{1} \and
Yue Yu\inst{1} \and
Bingzhuo Zhong\inst{1}\thanks{Corresponding author.}
}

\authorrunning{T. Yan et al.}

\institute{
The Hong Kong University of Science and Technology (Guangzhou)\\
\email{tyan497@connect.hkust-gz.edu.cn, bingzhuoz@hkust-gz.edu.cn}
}

\maketitle

\begin{abstract}
 Industrial deployment of robotic visual anomaly detection (VAD) is fundamentally constrained by passive perception under diverse 6-DoF pose configurations and unstable operating conditions such as illumination changes and shadows, where intrinsic semantic anomalies and physical disturbances coexist and interact. To overcome these limitations, a paradigm shift from passive feature learning to Active Canonicalization is proposed. \textit{\textbf{PiCo}} (\textbf{P}ose-\textbf{i}n-\textbf{Co}ndition Canonicalization) is introduced as a unified framework that actively projects observations onto a condition-invariant canonical manifold. \textit{\textbf{PiCo}} operates through a cascaded mechanism. The first stage, Active Physical Canonicalization, enables a robotic agent to reorient objects in order to reduce geometric uncertainty at its source. The second stage, Neural Latent Canonicalization, adopts a three-stage denoising hierarchy consisting of Photometric processing at the input level, Latent refinement at the feature level, and Contextual reasoning at the semantic level, progressively eliminating nuisance factors across representational scales. Extensive evaluations on the large-scale M$^2$AD benchmark demonstrate the superiority of this paradigm. \textit{\textbf{PiCo}} achieves a state-of-the-art 93.7\% O-AUROC, representing a 3.7\% improvement over prior methods in static settings, and attains 98.5\% accuracy in active closed-loop scenarios. These results demonstrate that active manifold canonicalization is critical for robust embodied perception.
\end{abstract}

\section{Introduction}
\label{sec:introduction}
\input{1_Intro}

\section{Related Works}
\label{sec:related works}

\input{2_RW}

\section{Methodology}
\label{sec:methodology}
\input{3_Method}

\section{Experiments and Experience}
\label{sec:experiments}
\input{4_Exp}

\section{Conclusion}
\label{sec:conclusion}
\input{5_Con}

\par\vfill\par
% Now we have reached the maximum length of an ECCV \ECCVyear{} submission (excluding references and acknowledgements).
% References should start immediately after the main text, but can continue past p.\ 14 if needed. 
\clearpage  % TODO FINAL: This \clearpage needs to be removed from both review and camera-ready versions.

%\section*{Acknowledgements}
% Please insert your acknowledgments here.


\bibliographystyle{unsrt}
\bibliography{main}
\end{document}

%% file: 1_Intro.tex
The rapid evolution of embodied AI is reshaping modern manufacturing, transitioning quality inspection from static, camera-based monitoring to dynamic, robot-centric interaction. Embodied industrial perception tasks agents with identifying rare and subtle defects on complex 3D objects in unstructured environments. While recent unsupervised learning paradigms have achieved nearly saturated performance on static computer vision benchmarks (e.g., MVTec AD \cite{bergmann2019mvtec}), their reliability collapses when deployed on general-purpose robots \cite{cao2024m2ad,wang2024realiad}. The root cause is not merely a domain gap, but a fundamental scientific disconnect between passive learning assumptions and active physical realities. 

\begin{figure}[!htbp]
  \centering
  \includegraphics[trim=2cm 6cm 2cm 3cm, clip, width=\textwidth]{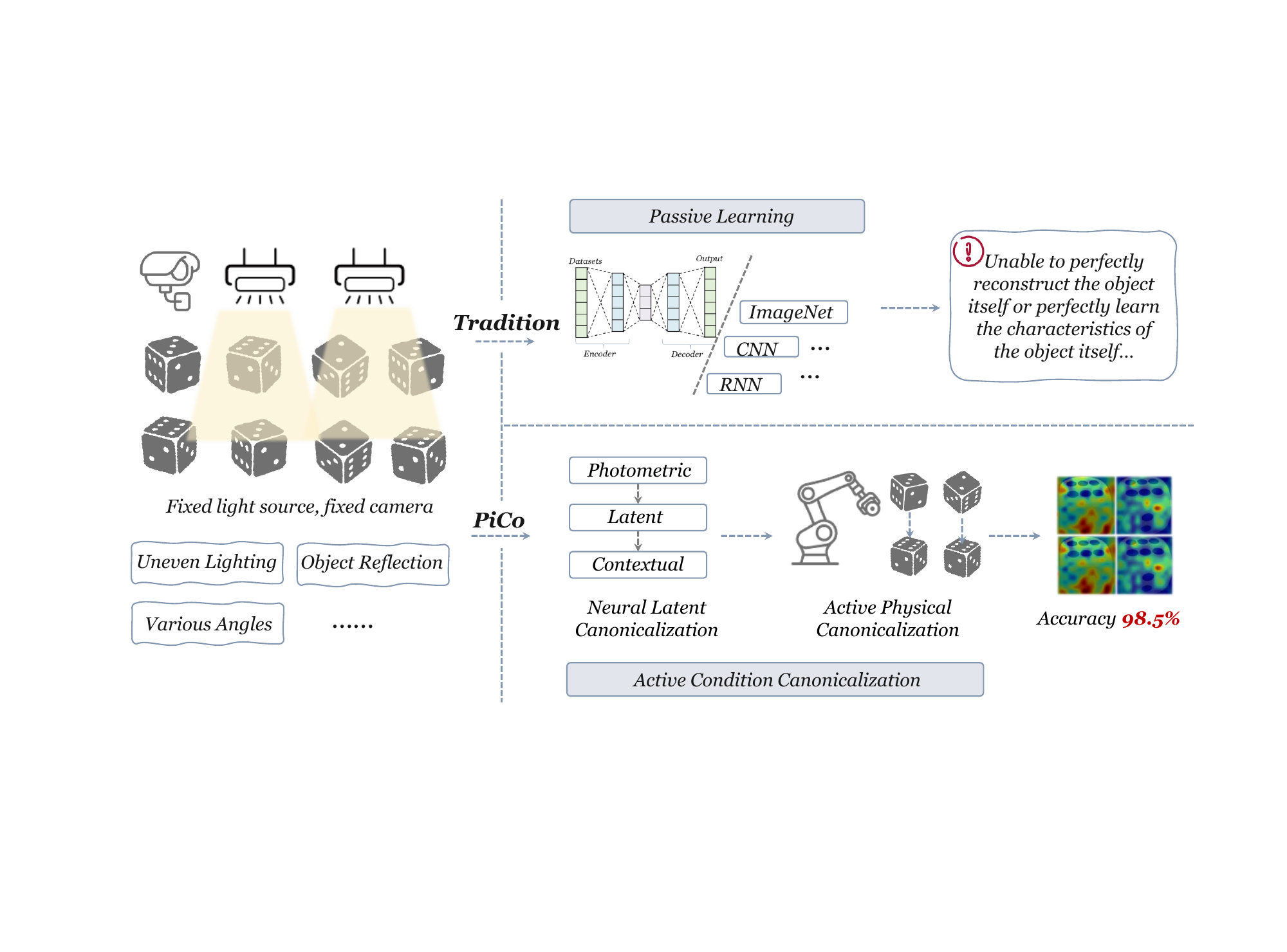}
  \caption{\textbf{Passive Learning vs Active Canonicalization: PiCo Framework for VAD.} Conventional visual anomaly detection relies on passive feature learning under fixed viewpoints and illumination, resulting in limited robustness to pose variations, uneven lighting, and specular artifacts. In contrast, PiCo integrates Active Physical Canonicalization and Neural Latent Canonicalization to actively reduce condition-induced variability, achieving 98.5\% accuracy under closed-loop operation.}
  \label{fig:cropped}
\end{figure}

% In static datasets, imaging conditions act as a constant bias. In contrast, within a robotic workcell, the observed visual signal $\mathbf{x}$ is a stochastic function jointly determined by the coupled Robot–Object pose configuration in 6-DoF and the surrounding environmental field, including illumination and occlusion. This phenomenon is defined as \textit{Semantic–Nuisance Entanglement}. From an information-theoretic perspective, the visual data stream is dominated by the high entropy of extrinsic nuisance variables denoted as condition $\mathbf{c}$, which obscures the sparse and low-entropy signal of intrinsic semantic defects denoted as state $\mathbf{s}$. Standard reconstruction-based autoencoders are optimized to maximize the mutual information between the input and the latent representation, expressed as $I(\mathbf{x}; \mathbf{z})$. In dynamic environments where $H(\mathbf{c}) \gg H(\mathbf{s})$, this objective implicitly biases the model toward encoding high-variance pose and illumination fluctuations, thereby reconstructing environmental variations with the same fidelity as structural defects. This effect is characterized as Nuisance Overfitting and represents a fundamental limitation of passive perception. A condition-invariant representation cannot be reliably learned from data distributions in which condition variance systematically dominates semantic variance. 

\textbf{The Scientific Challenge: Semantic-Nuisance Entanglement.} In static datasets, imaging conditions act as a constant bias. But within robotic contexts, the visual signal $\mathbf{x}$ is a stochastic function jointly determined by 6-DoF pose configuration and environmental factors such as illumination and occlusion. This phenomenon, termed \textit{Semantic–Nuisance Entanglement}, implies that the visual stream is dominated by high-entropy extrinsic variables $\mathbf{c}$ and low-entropy intrinsic defect state $\mathbf{s}$. Standard reconstruction-based autoencoders maximize $I(\mathbf{x}; \mathbf{z})$, encouraging the latent representation to preserve dominant variations in the input. When $H(\mathbf{c}) \gg H(\mathbf{s})$, this objective biases the model toward encoding pose and illumination fluctuations rather than semantic defects, leading to condition variance systematically dominating semantic variance.

\textbf{Paradigm Shift: Passive Adaptation to Active Canonicalization.} Existing approaches attempt to use large-scale data augmentation with the expectation that invariance can emerge from sufficient exposure to distributional variability. However, such strategies are computationally inefficient and theoretically suboptimal \cite{chen2020simclr,he2020moco}, without striking lance between nuisance and semantics. A paradigm shift toward \textit{Active Canonicalization} is therefore proposed. Robustness is reframed not solely as a representation learning problem, but as a joint perception-control problem \cite{ma2024canonicalization}. Rather than passively modeling high-entropy observations, the agent actively intervenes to contract the distribution onto a low-entropy canonical manifold. Based on this principle, Pose-in-Condition Canonicalization \textit{\textbf{PiCo}} is introduced as a unified framework integrated with Active Exploration and Representation Disentanglement:

\begin{enumerate}
    \item \textbf{Active Physical Canonicalization:} Rather than passively processing stoc-hastic observations, the robotic agent leverages epistemic uncertainty as a control feedback signal and actively reorients the object toward a canonical view. This geometric configuration is defined as the pose that minimizes ambiguity between superficial surface artifacts and intrinsic structural defects, thereby reducing pose-induced uncertainty at the physical level.
    
    \item \textbf{Neural Latent Canonicalization:} To suppress residual variability, a  Condition Invariant Bottleneck (CIB) mechanism is incorporated into the representation learning process. Realized through a three-stage denoising hierarchy consisting of Photometric processing at the signal level, Latent refinement at the feature level, and Contextual reasoning at the semantic level, this module imposes explicit information constraints that progressively attenuate nuisance factors across representational scales.
\end{enumerate}

The contributions are summarized as follows:

\begin{itemize}[label=$\bullet$]
    \item \textbf{Theoretical Formulation:} Robotic visual anomaly detection is reformulated as a Nuisance Disentanglement problem under entropy imbalance. It is shown that Active Canonicalization establishes a tractable upper bound on defect detectability in high-entropy observation regimes by explicitly reducing condition-induced uncertainty.
    
    \item \textbf{Unified Architecture:} The proposed PiCo unifies physics-guided photometric normalization, spectral filtering bottlenecks, and global linear attention within a coherent CIB framework, enabling structured suppression of nuisance variability across representational levels.
    
    \item \textbf{Empirical Breakthrough:} Extensive evaluation on $M^2{AD}$ and the robotic PiCo-Bench demonstrates state-of-the-art performance, achieving 93.7\% O-AUROC under static settings and 98.5\% accuracy in closed-loop active scenarios. These results substantiate that active agency fundamentally mitigates the brittleness inherent in passive visual perception.
\end{itemize}

\begin{figure}[htbp]
  \centering
  \includegraphics[trim=1cm 5cm 2cm 5cm, clip, width=\textwidth]{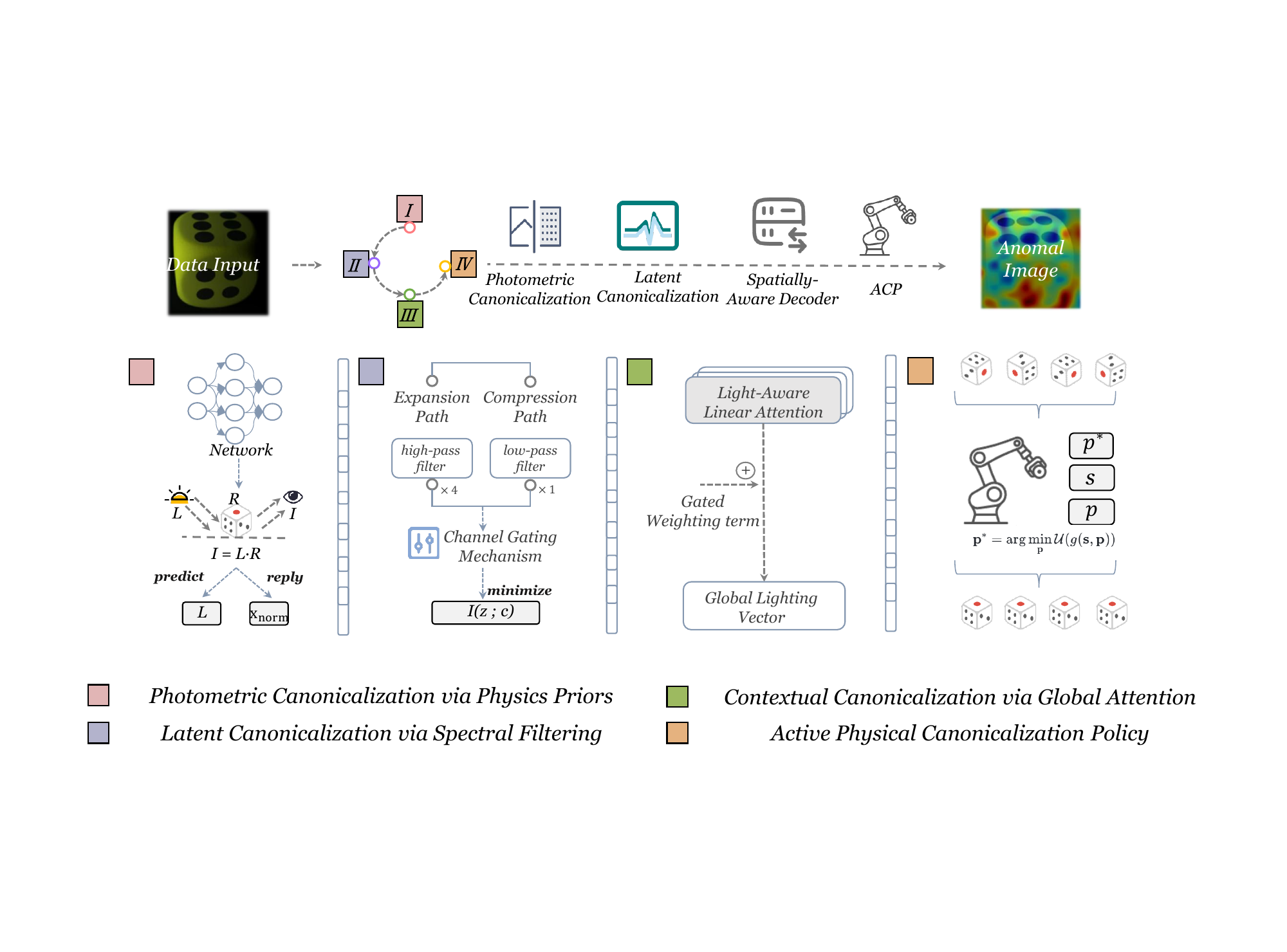}
  \caption{\textbf{Multi-stage Canonicalization Pipeline of the PiCo Framework.} This cascaded architecture progressively disentangles semantics from physical nuisances across three neural levels: photometric (Stage I), latent spectral (Stage II), and global contextual (Stage III). The resulting invariant representations ultimately drive the active physical canonicalization policy.}
  \label{fig:cropped}
\end{figure}

%% file: 2_RW.tex
\subsection{Unsupervised Visual Anomaly Detection}

Unsupervised visual anomaly detection has progressed from early reconstruction-based frameworks \cite{bergmann2019mvtec,zou2022visa} to modern feature-embedding paradigms. Recent state-of-the-art methods predominantly leverage pre-trained representations, often derived from ImageNet, to model normality in feature space. Memory-based approaches such as PatchCore \cite{roth2022patchcore} and CFA \cite{lee2022cfa} store nominal embeddings for nearest-neighbor inference, achieving strong performance on well-aligned benchmarks including MVTec AD \cite{bergmann2019mvtec} and VisA \cite{zou2022visa}.

Recent work has further emphasized efficiency and scalability. SimpleNet \cite{liu2023simplenet} and EfficientAD \cite{batzner2024efficientad} reduce computational overhead via lightweight adaptation and student–teacher distillation, while RD++ \cite{tran2023rdpp} and MSFlow \cite{zhang2024msflow} improve accuracy through refined reverse distillation and multi-scale normalizing flows. The field is also shifting toward universal and multi-class formulations \cite{tolstikhin2021mlpmixer,guo2022hiremlp}, where methods such as Dinomaly \cite{guo2025dinomaly} and INP-Former \cite{luo2025inpformer} show that foundation features and intrinsic normal prototypes can generalize across categories without class-specific training.

Despite these advances, existing approaches remain passive, assuming near-canonical poses and controlled illumination. Robotic-centric benchmarks such as Real-IAD \cite{wang2024realiad} and M$^2$AD \cite{cao2024m2ad} show significant degradation under dynamic pose–illumination interactions. Diffusion-based methods have been explored to model such variability, e.g., RSDNet \cite{qu2026rsdnet}, which employs detachable latent diffusion for efficient single-step inference and robustness to multi-level noise. In contrast, \textit{\textbf{PiCo}} introduces an active canonicalization mechanism to reduce condition-induced variability at its source.

\subsection{Invariance Learning and Disentanglement}

Learning representations invariant to nuisance variability has long been a central goal in computer vision \cite{alemi2017deep}. Contrastive frameworks such as SimCLR \cite{chen2020simclr} and MoCo \cite{he2020moco} promote semantic robustness by aligning features across augmented views. In visual anomaly detection, methods like CutPaste \cite{li2021cutpaste} and DRAEM \cite{zavrtanik2021draem} employ synthetic augmentations (e.g., region replacement, photometric perturbations) to encourage insensitivity to defect-like artifacts. From a theoretical perspective, the Information Bottleneck \cite{tishby2000ib} and its conditional variant \cite{fischer2020cib} provide a principled framework for learning minimal sufficient representations by compressing nuisance information \cite{lyu2023recognizable} while preserving task-relevant semantics.

However, existing VAD methods model nuisance variability via low-dimensional image-space transformations, failing to capture complex effects in real robotic environments. In contrast, \textit{\textbf{PiCo}} uses embodied physical augmentation, where robot-induced pose variations form positive pairs under a Conditional Information Bottleneck, yielding invariance within the robot’s operating envelope. Diffusion-inspired feature perturbations further improve robustness to such variability \cite{qu2025cdsegnet}.

Related advances in unsupervised representation learning and clustering provide complementary insights. CNMBI~\cite{zhang2023cnmbi} dynamically determines cluster numbers via intrinsic data distributions while filtering low-confidence samples, and IDCL leverages density-guided curriculum learning with density cores to improve clustering stability and convergence on complex datasets \cite{zhang2024idcl}.

% \subsection{Active Perception in Robotics}

% Active perception, originally formalized by Bajcsy \cite{bajcsy1988active}, enables robotic systems to deliberately control sensing configurations in order to improve information acquisition. A major research direction within this domain is Next-Best-View (NBV) planning \cite{daudelin2017nbv}, which optimizes camera viewpoints for objectives such as three-dimensional reconstruction or grasp coverage maximization.

% Active perception has been widely studied in recognition and SLAM, but its use in active anomaly detection remains underexplored \cite{chamiti2025lmad,lutzelberger2023active}. Conventional NBV methods maximize information gain or surface coverage, which may favor viewpoints with specular highlights or occlusions that obscure subtle defects. \textit{\textbf{PiCo}} overcomes this via an Active Canonicalization Policy that directly maximizes the anomaly score’s signal-to-noise ratio. By manipulating the object toward configurations that reduce Semantic–Nuisance Entanglement, the policy minimizes condition-induced uncertainty and improves defect discriminability.

\subsection{Active Perception in Robotics}

Active perception, originally formalized by Bajcsy \cite{bajcsy1988active}, enables robots to control sensing configurations for improved information acquisition. A key direction is Next-Best-View (NBV) planning \cite{daudelin2017nbv}, which optimizes viewpoints for tasks such as 3D reconstruction and grasp coverage.

While widely studied in recognition and SLAM, its application to anomaly detection remains limited \cite{chamiti2025lmad,lutzelberger2023active}. \textit{\textbf{PiCo}} addresses this with an Active Canonicalization Policy that directly maximizes the anomaly score’s signal-to-noise ratio. By steering objects toward configurations that reduce Semantic–Nuisance Entanglement, the policy minimizes condition-induced uncertainty and improves defect discriminability.

%% file: 3_Method.tex
\textit{\textbf{PiCo}} is a robotic visual perception framework specifically designed to address the complexities inherent in robotic vision, including pose-dependent camera viewpoints and dynamically varying environmental conditions such as illumination changes. The architecture comprises three principal components: a pre-trained Vision Transformer encoder augmented with an illumination preprocessing module, a dual-structured multilayer perceptron bottleneck for structured feature filtering, and a decoder equipped with attention mechanisms that incorporate global posture-aware context. 

In contrast to conventional neural anomaly detection models, \textit{\textbf{PiCo}} explicitly emphasizes pose-aware representation learning and canonicalization. By systematically modeling positional and environmental variability, the framework preserves defect-relevant semantics while suppressing condition-induced noise, thereby enhancing both accuracy and robustness under diverse robotic operating conditions.

Fundamentally, the effectiveness of \textit{\textbf{PiCo}} stems from its structured integration of canonicalization at both the physical and representational levels. By leveraging the strong semantic prior of a pre-trained Vision Transformer (ViT), the framework systematically suppresses extrinsic environmental nuisances while preserving intrinsic defect features, ensuring robust generalization across diverse operating conditions.

\begin{figure}[!htbp]
  \centering
  \includegraphics[trim=3cm 2cm 3cm 4cm, clip, width=\textwidth]{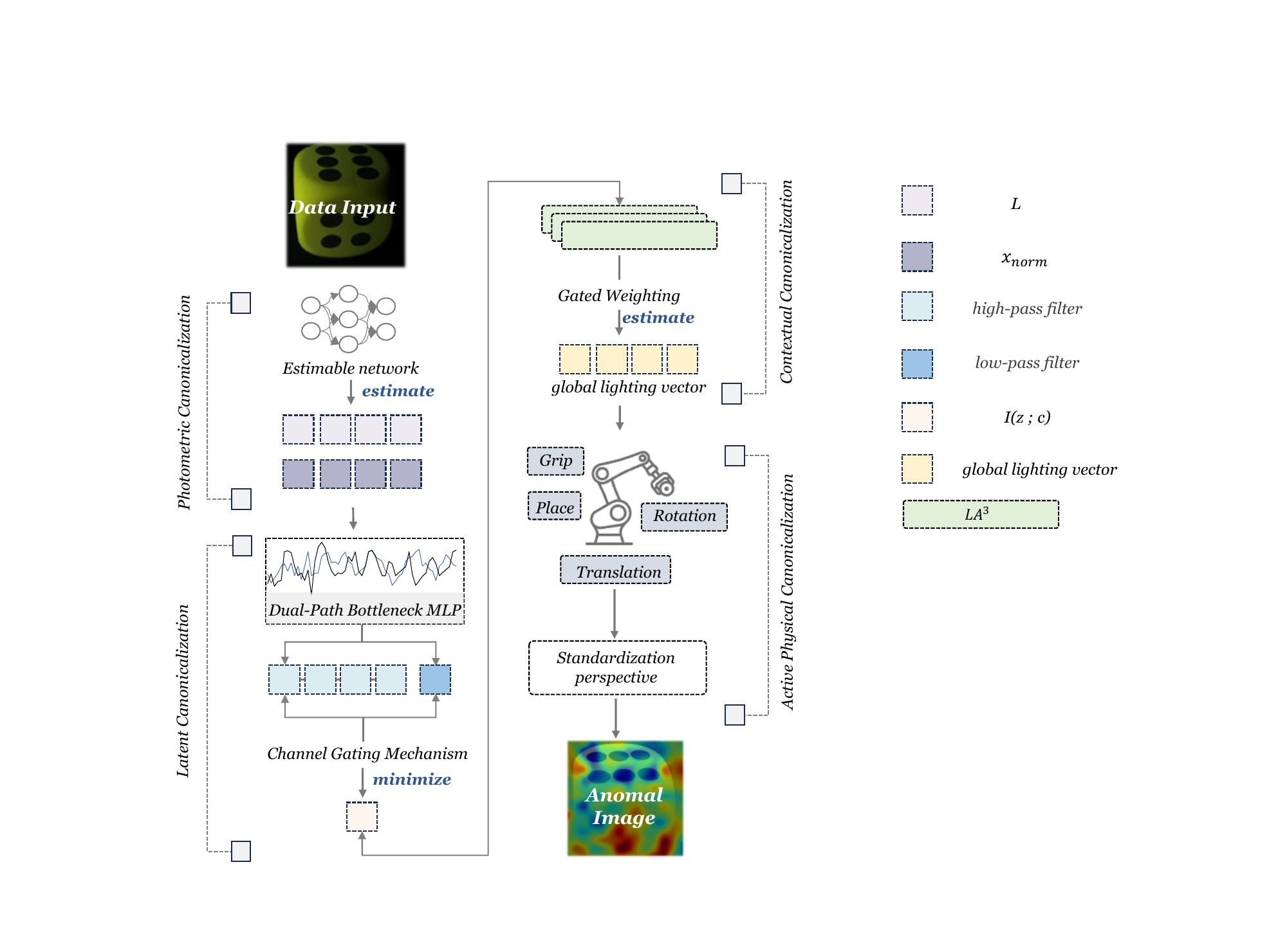}
  \caption{\textbf{Modular Architecture of PiCo's Canonicalization Mechanism.} The framework integrates a photometric preprocessing module for lighting estimation, a dual-path bottleneck MLP for latent feature refinement, and contextual canonicalization. Ultimately, reconstruction uncertainty serves as a real-time feedback signal, driving the robotic agent to actively seek a geometrically optimal canonical view.}
  \label{fig:3}
\end{figure}

\subsection{Preliminary Knowledge}
\label{subsec:pre}
The problem is formalized from an information-theoretic perspective. Let an observation $\mathbf{x} \in \mathcal{X}$ be generated by two latent factors: intrinsic semantic state $\mathbf{s}$ and extrinsic condition variable $\mathbf{c}$, where $\mathbf{c}$ encompasses pose and illumination variations.
Standard VAD maximizes the mutual information $I(\mathbf{x}; \mathbf{z})$ for reconstruction. However, since $H(\mathbf{c}) \gg H(\mathbf{s})$ in dynamic environments, maximizing this term inevitably maximizes $I(\mathbf{c}; \mathbf{z})$, leading to nuisance overfitting.
Theorem 1 (Robustness via Conditional Information Bottleneck).
To achieve robust detection, the representation $\mathbf{z}$ must satisfy the CIB objective:

\begin{equation}
    \min_{\theta} \mathcal{L}_{CIB} = \underbrace{-I(\mathbf{z}; \mathbf{s})}_{\text{Maximizing Semantics}} + \beta \underbrace{I(\mathbf{z}; \mathbf{c})}_{\text{Minimizing Nuisance}}
\end{equation}
PiCo approximates this objective via a Cascaded Canonicalization mechanism, progressively minimizing $I(\mathbf{z}; \mathbf{c})$ at Photometric, Latent, and Contextual levels.

\subsection{Stage I: Environmental Normalization via Illumination-preprocessing}
\label{subsec:stage_i}

Deep visual perception models often show significant performance degradation under non-ideal lighting conditions, which are prevalent in industrial and robotic contexts. To address this issue, \textit{\textbf{PiCo}} incorporates a plug-and-play illumination preprocessing module. Inspired by the DI-Retinex framework \cite{sun2024di}, the observed image $\mathbf{I}$, corresponding to the observation variable $\mathbf{x}$, is modeled as the element-wise Hadamard product of intrinsic reflectance $\mathbf{R}$ and the spatial illumination field $\mathbf{L}$: 
\begin{equation}
    \mathbf{I} = \mathbf{R} \circ \mathbf{L}
\end{equation}
To minimize the variance caused by photometric interference (part of c) at the input end, we introduce a differentiable estimation network  $f_{\phi}(\cdot)$ to predict the non-uniform continuous illumination field. This network explicitly simulates the quantization error and dynamic range overflow of digital images. Subsequently, we restore the normalized reflectance map $\mathbf{x}_{norm}$ through element-wise division:
\begin{equation}
    \mathbf{x}_{norm} = \mathbf{I} \oslash \big(f_{\phi}(\mathbf{I}) + \epsilon\big)
\end{equation}
where $\oslash$ denotes the element-wise division operator, and $\epsilon$ is the numerical stability term.

By explicitly correcting non-uniform lighting and shadow artifacts at the raw signal level, this module provides the subsequent Vision Transformer (ViT) encoder with a robust, illumination-invariant reflectance map ($\mathbf{x}_{norm}$). This front-end mitigation effectively prevents photometric distortions from propagating into the feature space, ensuring highly consistent representations even under the extreme lighting fluctuations prevalent in robotic workcells.

\subsection{Stage II: Latent Canonicalization via Spectral Filtering}

While photometric calibration normalizes illumination, pose-induced geometric artifacts persist in the feature space. To explicitly enforce the Conditional Information Bottleneck (CIB) goal by minimizing $I(\mathbf{z}; \mathbf{c})$, we introduce a conditionally invariant bottleneck instantiated as a Dual-Path Spectral Filtering MLP (bMLP). Typically, intrinsic semantic defects $\mathbf{s}$ manifest as isolated high-frequency anomalies, whereas pose variations $\mathbf{c}$ induce low-rank structural shifts or widespread noise. To effectively decouple these in the frequency domain, our bMLP employs two parallel branches:

\begin{itemize}[label=$\bullet$] 
    \item \textbf{Expansion path ($\mathcal{F}_{high}$):} Expands the channel dimension by $4\times$ to act as a high-pass filter, enabling the lossless retention of fine-grained defect details.
    \item \textbf{Compression path ($\mathcal{F}_{low}$):} Compresses features to act as a low-rank information bottleneck (low-pass filter), explicitly discarding redundant pose-related nuisances.
\end{itemize}

For input features $\mathbf{z}_{in}$, adaptive band selection is achieved via a channel gating mechanism $\mathbf{g} \in [0, 1]^C$, driven by 1D convolution and adaptive average pooling:
\begin{equation}
    \mathbf{g} = \sigma\Big(\text{Conv1D}\big(\text{AvgPool}(\mathbf{z}_{in})\big)\Big)
\end{equation}

Rather than standard concatenation, the final canonicalized feature $\mathbf{z}_{out}$ is aggregated using a highly efficient additive gated fusion:
\begin{equation}
    \mathbf{z}_{out} = \text{PreNorm}\Big(\mathbf{g} \odot \mathcal{F}_{high}(\mathbf{z}_{in}) + (\mathbf{1} - \mathbf{g}) \odot \mathcal{F}_{low}(\mathbf{z}_{in})\Big)
\end{equation}

Augmented with PreNorm and DropPath regularization to stabilize gradient propagation, this gating mechanism dynamically suppresses channels heavily correlated with transient pose interference. By doing so, the architecture topologically approximates the optimal CIB solution, stripping away physical nuisances while rigorously retaining intrinsic defect semantics.

\subsection{Stage III: Contextual Canonicalization via Spatial-Aware Decoder}

Standard ViT decoders relying on softmax-based self-attention are computationally expensive and prone to overfitting local high-contrast edges, such as shadow boundaries. To address this, we design an 8-layer spatial-aware decoder driven by a Light-Aware Linear Attention ($\text{LA}_3$) mechanism equipped with Channel Attention Gating (CAG). 

Given queries $\mathbf{Q}$, keys $\mathbf{K}$, and values $\mathbf{V} \in \mathbb{R}^{N \times d}$, we remove the softmax operator and approximate the attention matrix using a positive kernel feature map $\phi(x) = \text{ELU}(x) + 1$. This reduces the computational complexity to linear $\mathcal{O}(N)$. The $\text{LA}_3$ operation is formulated as:
\begin{equation}
    \text{Attention}(\mathbf{Q}, \mathbf{K}, \mathbf{V}) = \text{Clamp}\left( \frac{\phi(\mathbf{Q})(\phi(\mathbf{K})^T\mathbf{V})}{\phi(\mathbf{Q})(\phi(\mathbf{K})^T\mathbf{1}) + \epsilon} \right) \odot \mathbf{W}_{CAG}
\end{equation}
where $\mathbf{1}$ is an all-ones vector, and $\epsilon$ alongside the $\text{Clamp}(\cdot)$ function ensures numerical stability against gradient explosion during optimization. $\mathbf{W}_{CAG}$ represents the adaptive channel gating weights. 

By aggregating global context linearly, $\text{LA}_3$ expands the effective receptive field to implicitly model the scene-level illumination distribution. This contextual canonicalization enables the decoder to reliably disambiguate dynamic shadow deformations from genuine structural defects, maximizing the specificity of anomaly localization.

\begin{algorithm}[!htbp]  
\caption{PiCo: Active Manifold Canonicalization Pipeline}
\label{alg:pico_pipeline}
\small 
\begin{algorithmic}[1]
\REQUIRE $I_0$ (initial observation), $\mathbf{p}_0$ (initial 6-DoF pose), $T$ (max steps), $\tau$ (uncertainty threshold), $\mathcal{F}_{photo}$, $\mathcal{E}$, $\mathcal{B}$, $\mathcal{D}$
\STATE $t \leftarrow 0$, $\text{Found} \leftarrow \text{False}$ \COMMENT{Init flags}

\WHILE{$t \leq T$ \AND $\neg\text{Found}$}
    \STATE $\mathbf{x}_{norm}^{(t)} \leftarrow \mathcal{F}_{photo}(I_t)$ \COMMENT{Stage I: Photometric Canonicalization}
    \STATE $\mathbf{z}_{in}^{(t)} \leftarrow \mathcal{E}(\mathbf{x}_{norm}^{(t)})$ \COMMENT{Stage II: Latent Canonicalization}
    \STATE $\mathbf{z}_{out}^{(t)} \leftarrow \mathcal{B}(\mathbf{z}_{in}^{(t)})$
    \STATE $\hat{\mathbf{x}}^{(t)} \leftarrow \mathcal{D}(\mathbf{z}_{out}^{(t)})$ \COMMENT{Stage III: Contextual Canonicalization}
    
    \STATE $\mathcal{U}_t \leftarrow \text{Calculate\_Uncertainty}(\mathbf{x}_{norm}^{(t)}, \hat{\mathbf{x}}^{(t)})$ \COMMENT{Active Policy}
    \IF{$\mathcal{U}_t < \tau$}
        \STATE $\text{Found} \leftarrow \text{True}$
        \STATE $\mathcal{M}_{final} \leftarrow \text{Compute\_Anomaly\_Map}(\mathbf{x}_{norm}^{(t)}, \hat{\mathbf{x}}^{(t)})$
    \ELSE
        \STATE $\mathbf{p}_{t+1} \leftarrow \arg\min_{\mathbf{p}} \mathbb{E}\left[\mathcal{U}(\mathbf{p})\right]$
        \STATE $I_{t+1} \leftarrow \text{Robot\_Execute}(\mathbf{p}_{t+1})$, $t \leftarrow t + 1$
    \ENDIF
\ENDWHILE

\RETURN $\mathcal{M}_{final}$ (Final Anomaly Map)
\end{algorithmic}
\end{algorithm}
\vspace{-8pt}

\subsection{Active Canonicalization Policy}
\label{subsec:ACP}

% While the neural canonicalization stages operate at the representational level to suppress intrinsic feature noise, they remain fundamentally constrained by the quality of the originally captured observation. To overcome the limitations of fixed-view perception, including occlusion, severe perspective distortion, and specular highlights, PiCo incorporates an Active Canonicalization Policy. Under this policy, the robotic system actively adjusts its 6-DoF pose relative to the object, seeking a geometrically favorable configuration referred to as the canonical view. This configuration is defined as the pose at which aleatoric uncertainty induced by environmental and geometric factors is minimized, and the separability between normal and anomalous patterns is maximized. By intervening at the physical acquisition stage rather than solely at the feature level, Active Canonicalization reduces condition-induced ambiguity at its source, thereby complementing neural denoising with control-driven uncertainty suppression. 

Although neural canonicalization effectively suppresses feature-level noise, it remains fundamentally constrained by the physical quality of the initial observation. To address this limitation, \textit{\textbf{PiCo}} introduces an Active Canonicalization Policy that drives the robot to actively adjust its 6-DoF pose in search of a canonical view—defined as a geometric configuration where environmental uncertainty is minimized and the separability between normal and anomalous patterns is maximized. 

This physical intervention reduces environment-induced ambiguity at the source, thereby complementing the limitations of purely neural denoising. Using the divergence between the encoded latent projection and the decoded reconstruction, we rigorously define the epistemic uncertainty $\mathcal{U}$ of the model as the squared $L_2$-norm of the pixel-level reconstruction error:
\begin{equation}
    \mathcal{U}(g(\mathbf{s}, \mathbf{p})) = \|\mathbf{x}_{norm} - \hat{\mathbf{x}}\|_2^2
\end{equation}

When $\mathcal{U}$ exceeds a preset confidence threshold $\tau$, the system defers the anomaly decision for the current frame. Instead, it utilizes $\mathcal{U}$ as a real-time feedback signal to trigger an active repositioning policy, seeking the optimal canonical view $\mathbf{p}^*$:
\begin{equation}
    \mathbf{p}^* = \arg\min_{\mathbf{p}} \mathcal{U}(g(\mathbf{s}, \mathbf{p}))
\end{equation}

This control-driven formulation actively navigates intractable out-of-distribu-tion (OOD) observations in the physical workspace toward a high-likelihood manifold region learned by the model, effectively resolving the physical entanglement between environmental nuisances (e.g., shadows, specular glare) and intrinsic semantic defects.

%% file: 4_Exp.tex
The experimental protocol is designed to systematically investigate three fundamental aspects of the proposed cascaded canonicalization paradigm:

\textbf{Q1 (Static Superiority).} Evaluation of whether the neural canonicalization hierarchy enables PiCo to surpass state-of-the-art passive methods on the challenging M$^2$AD benchmark.

\textbf{Q2 (Component Efficacy).} Analysis of the individual and collective contributions of the Photometric, Latent, and Contextual canonicalization stages to overall robustness.

\textbf{Q3 (Active Capability).} Assessment of whether the active physical canonicalization policy can resolve failure cases that remain intractable under purely static perception.

\subsection{Experimental Setup}
\label{subsec:setup}

\subsubsection{Datasets}

\begin{itemize}[label=$\bullet$]
    \item \textbf{M\textsuperscript{2}AD} \textbf{(Static Protocol).} The full-scale M$^2$AD dataset is adopted for evaluation, comprising 119,000 high-resolution images spanning 10 industrial object categories. Each object instance is captured under 120 distinct view and illumination configurations, providing a comprehensive distribution of pose and photometric variations. This dataset therefore serves as the primary benchmark for assessing static robustness under diverse geometric and environmental conditions.

  \begin{figure}[!htbp]
    \centering
     \includegraphics[width=0.9\linewidth]{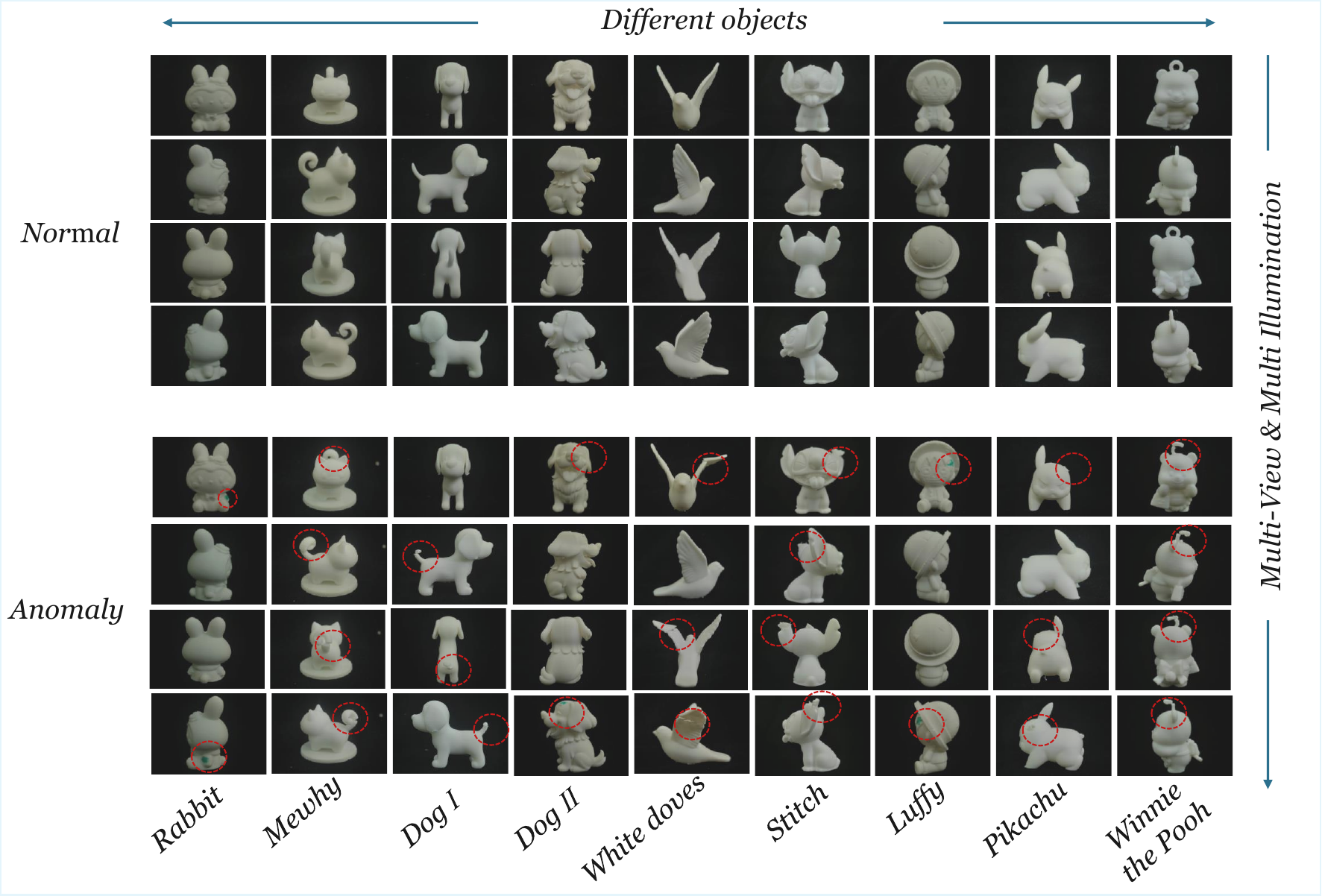}    \caption{\textbf{Representative cases from the geometrically complex subset of \textit{PiCo-Bench.}} Each block compares normal and anomalous samples, with subtle structural defects marked by red dotted circles. Their strong non-planarity induces pose-dependent \& self-occlusion, exemplifying semantic–nuisance entanglement.}
    \label{fig:anomaly}
\end{figure}

    \item \textbf{PiCo-Bench (Dynamic Protocol).} To systematically evaluate closed-loop active perception, we introduce PiCo-Bench, a novel dataset curated within a real-world robotic workstation. As illustrated in Fig. \ref{fig:anomaly}, the dataset features complex 3D objects with pronounced non-planar topologies (e.g., intricate figurines) that host subtle, viewpoint-sensitive structural defects. Crucially, their intricate geometries naturally induce severe self-occlusions and dynamic shadows under varying 6-DoF poses, explicitly amplifying the Semantic-Nuisance Entanglement. Consequently, conventional static detection methods catastrophically fail under these conditions, underscoring the necessity of active manifold canonicalization for embodied robustness.

\end{itemize}

% The anomalies correspond to subtle structural defects that are highly sensitive to viewpoint variations, including missing components and localized surface damage. Owing to their complex topology, these objects frequently produce severe self-occlusions and dynamic self-shadowing under diverse 6-DoF pose configurations. This physical characteristic explicitly amplifies the problem of Semantic–Nuisance Entanglement, thereby establishing a rigorous testbed for evaluating active canonicalization strategies. Under this benchmark setting, conventional static anomaly detection approaches typically suffer significant performance degradation or complete failure, further highlighting the necessity of active manifold canonicalization for robust embodied perception.

This paper compares PiCo against state-of-the-art unsupervised VAD methods, including Dinomaly \cite{guo2025dinomaly}, RD++ \cite{tran2023rdpp}, MSFlow \cite{zhang2024msflow}, and INP-Former \cite{luo2025inpformer}. All baselines are re-trained on M$^2$AD following their official protocols.

All models are trained under a standardized protocol to ensure fair comparison. An identical backbone network pretrained with ImageNet is adopted and remains frozen unless explicitly specified.
Input images are uniformly resized to $256 \times 256$. 
In accordance with the synthetic anomaly paradigm, a cosine annealing schedule is employed for learning rate decay and optimization is performed using AdamW.
Furthermore, to ensure fair comparability with existing unsupervised anomaly detection methods, we strictly avoid dataset-specific hyperparameter tuning, exclude architecture-specific optimization tricks, and employ identical training schedules for both the baselines and our proposed approach.

\subsection{Comparative Analysis on $M^2AD$ (Addressing Q1)}
\label{subsec: Comparative Analysis}

Object-level AUROC (O-AUROC), Pixel-level AUROC (P-AUROC), and Pixel-level AUPRO are reported in Table~\ref{tab:comparative_results}. PiCo establishes a new state-of-the-art performance across evaluation metrics, achieving substantial improvements over existing baselines.

\vspace{-15pt}
\begin{table}[H]
\centering
\small
\setlength{\tabcolsep}{4pt}
\renewcommand{\arraystretch}{1.15}

\resizebox{\linewidth}{!}{
\begin{tabular}{lcccccc}
\toprule
\textbf{Category} 
& \makecell{\textbf{Dinomaly \cite{guo2025dinomaly}} \\ \small \textbf{CVPR'2025}} 
& \makecell{\textbf{RD++ \cite{tran2023rdpp}} \\ \small \textbf{CVPR'2023}} 
& \makecell{\textbf{MSFlow \cite{zhang2024msflow}} \\ \small \textbf{TNNLS'2024}} 
& \makecell{\textbf{INP-Former \cite{luo2025inpformer}} \\ \small \textbf{CVPR'2025}} 
& \textbf{PiCo (Ours)} 
& \textbf{Improvement (vs Dino)} \\
\midrule

Bird   & 75.1/74.9/86.9 & \textbf{90.3}/70.2/79.8 & 85.0/62.0/71.4 & 80.0/67.2/84.1 & \textbf{80.7/76.7/90.9} & +5.6 / +1.8 / +4.0 \\
Car    & 86.7/75.1/78.3 & 85.0/68.2/75.6 & 67.9/55.9/67.4 & 58.1/53.9/72.1 & \textbf{89.3/84.6/93.0} & +2.6 / +9.5 / +14.7 \\
Cube   & 82.3/77.8/86.0 & 83.1/74.6/80.7 & 66.0/57.8/58.7 & 77.9/74.5/80.6 & \textbf{96.8/82.4/92.6} & +14.5 / +4.6 / +6.6 \\
Dice   & 98.1/93.0/85.7 & \textbf{98.4}/89.4/85.6 & 76.8/69.4/77.0 & 93.3/83.7/87.7 & \textbf{96.7/94.0/90.5} & -1.4 / +1.0 / +4.8 \\
Doll   & 74.4/72.6/89.0 & 66.8/65.9/85.4 & 56.4/55.1/68.9 & 72.5/73.7/85.8 & \textbf{81.6/78.6/90.9} & +7.2 / +6.0 / +1.9 \\
Motor  & 95.4/85.4/94.2 & 92.2/87.9/94.9 & 86.0/61.4/86.7 & 83.7/61.1/91.9 & \textbf{99.6/92.2/95.5} & +4.2 / +6.8 / +1.3 \\
Holder & 99.7/85.8/90.0 & 99.1/\textbf{87.8}/81.0 & 98.0/76.6/59.6 & 99.2/76.4/81.0 & \textbf{100/86.0/97.7} & +0.3 / +0.2 / +7.7 \\
Ring   & 91.2/87.3/77.8 & \textbf{95.5/90.9}/77.2 & 74.7/72.4/83.9 & 75.5/71.7/\textbf{91.4} & 94.2/82.2/88.5 & +3.0 / -5.1 / +10.7 \\
Teapot & 99.9/94.6/94.3 & 91.3/86.0/91.7 & 83.0/63.9/77.3 & 91.6/79.1/92.4 & \textbf{100/95.1/98.0} & +0.1 / +0.5 / +3.7 \\
Tube   & 97.2/83.3/77.0 & 92.1/81.2/\textbf{90.9} & 89.0/67.3/84.1 & 78.0/64.1/85.9 & \textbf{98.3/85.1/86.4} & +1.1 / +1.8 / +9.4 \\
\midrule
\textbf{Average} 
& 90.0/83.0/85.9 
& 89.4/80.2/84.3 
& 78.3/64.2/73.5 
& 81.0/70.5/85.3 
& \textbf{93.7/85.7/92.4} 
& \textbf{+3.7 / +2.7 / +6.5} \\
\bottomrule
\end{tabular}
}
\caption{
Quantitative comparison on M$^2$AD. PiCo significantly outperforms the previous SOTA (Dinomaly) by +3.7\%, +2.7\% and +6.5\% on average.
}
\label{tab:comparative_results}
\end{table}
\vspace{-20pt}

% \begin{itemize}[label=$\bullet$]
%     \item \textbf{Overall Robustness:} PiCo achieves a record 93.7\% mean O-AUROC, surpassing the strongest competing method, Dinomaly, by 3.7\%. This performance gap indicates that explicitly modeling nuisance variability through the Conditional Information Bottleneck framework provides a measurable advantage over purely passive feature matching strategies.
%     \item \textbf{Geometry-Nuisance Decoupling:} The most pronounced improvements are observed in geometrically complex categories such as Cube (+14.5\%) and Motor (+4.2\%). These objects undergo substantial appearance variation under changing illumination conditions, including pronounced shading effects on planar surfaces. Whereas Dinomaly exhibits difficulty in separating illumination-induced artifacts from genuine structural defects, PiCo’s latent canonicalization mechanism effectively disentangles geometric structure from photometric variation, resulting in significantly improved detection stability.
% \end{itemize}

\textbf{Overall Robustness.} PiCo achieves a record \textbf{93.7\%} average O-AUROC, surpassing the strongest competing method, Dinomaly, by \textbf{3.7\%}. This performance gap indicates that explicitly modeling nuisance variability through the Conditional Information Bottleneck framework provides a measurable advantage over purely passive feature matching strategies.

\textbf{Geometry-Nuisance Decoupling.} The most pronounced improvements are observed in geometrically complex categories such as Cube (+14.5\%) and Motor (+4.2\%). These objects show substantial appearance variation under various illumination conditions, including shading effects on planar surfaces. \textit{\textbf{PiCo}}’s latent canonicalization mechanism effectively disentangles geometric structure from photometric variation, resulting in improved detection stability.

\vspace{-20pt}

\begin{figure}[!htbp]
    \centering
    \includegraphics[ width=1.0\textwidth]{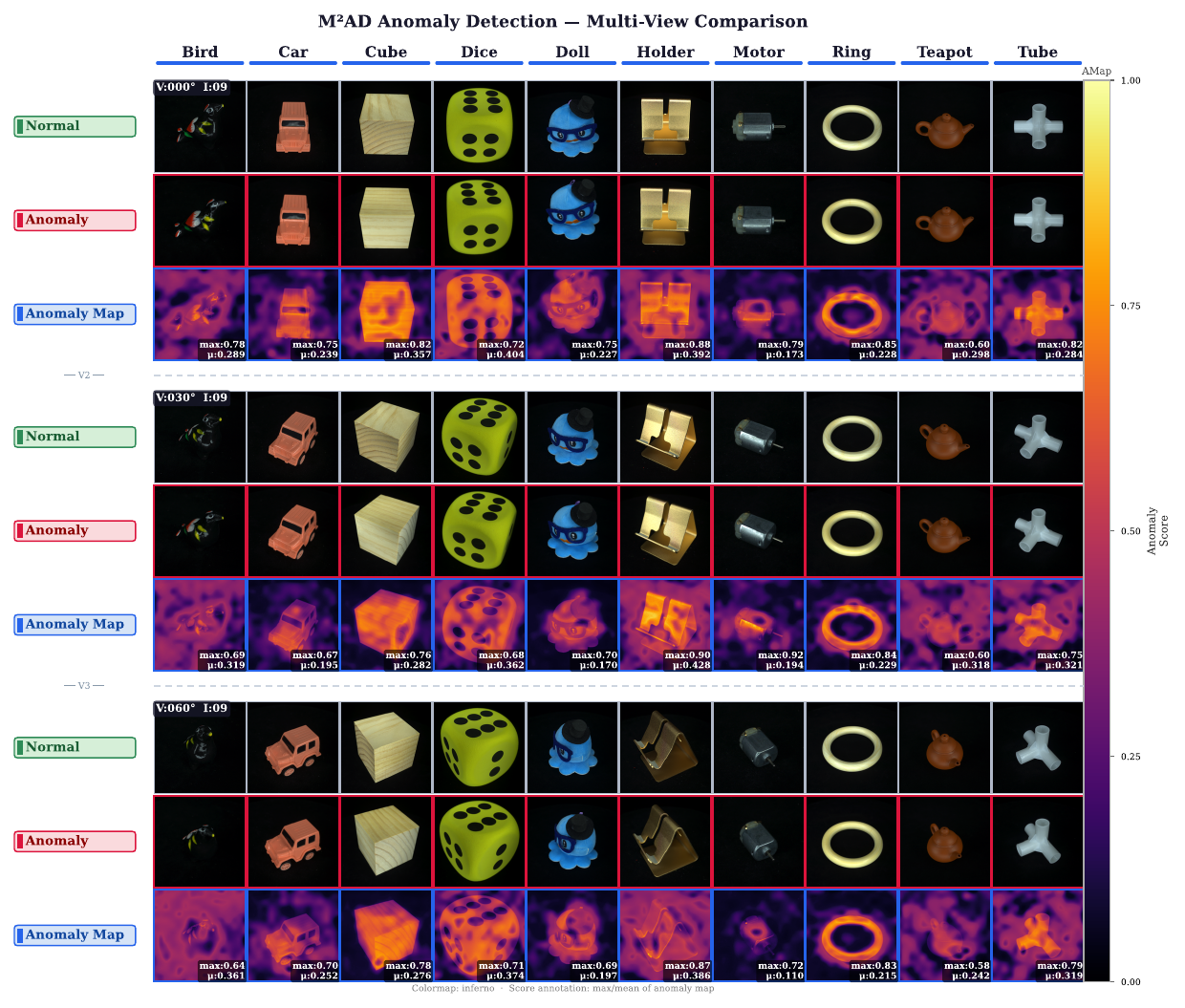}
    \caption{\textbf{Qualitative anomaly detection results on the $\mathbf{M^2AD}$ benchmark.} 
    The rows display normal image, anomalous image, PiCo's predicted anomaly map masked by ground truth. Evaluated under highly fluctuating illumination and viewpoints, PiCo precisely localizes semantic defects while suppressing environmental nuisances.}
    \label{fig:5}
\end{figure}

\vspace{-20pt}

\subsection{Ablation Study: Validating the Hierarchy (Addressing Q2)}
\label{subsec: Ablation Study}

To validate the Three-Stage Denoising hypothesis, a progressive ablation study is conducted in which each canonicalization stage is incrementally introduced. The corresponding results are summarized in Table~\ref{tab: ablation}.

\vspace{-15pt}
\begin{table}[H]
\centering
\small
\setlength{\tabcolsep}{4pt}
\renewcommand{\arraystretch}{1.15}

\resizebox{\linewidth}{!}{
\begin{tabular}{lccccccc}
    \toprule
    \textbf{Variant} & \textbf{bMlp} & \textbf{DIR ($\alpha=0.1$)} & \textbf{DIR ($\alpha=0.2$)} & \textbf{$LA_3$} & \textbf{AUROC-O} & \textbf{AUROC-I} & \textbf{AUPRO} \\
    \midrule
    (a) (Baseline) & \checkmark & & & & 93.33 & 86.51 & 91.77 \\
    (b) & \checkmark & \checkmark & & & 93.35 & 86.46 & 92.06 \\
    (c) & \checkmark & & \checkmark & & 93.00 & 86.19 & 91.97 \\
    \textbf{(d) (Full)} & \checkmark & \checkmark & & \checkmark & \textbf{93.72} & \textbf{85.68} & \textbf{92.40} \\
    \bottomrule
\end{tabular}
}
\caption{
\textbf{Ablation Study on Neural Components.} Dual-path bMlp with DIR($\alpha=0.1$) + Light-Aware Linear Attention shows optimal balance for manifold constraints.}
\label{tab: ablation}
\end{table}
\vspace{-20pt}

\textbf{Latent Canonicalization (Variant A).} Employing only the Dual-Path BMLP establishes a strong baseline of 93.33i\%. This result supports the hypothesis that spectral filtering through dual pathways, which separates high-frequency defect signals from low-rank nuisance components, provides inherent structural robustness.
    
\textbf{Manifold Constraints (Variant B vs. C).} Incorporating the DIR Loss with $\alpha = 0.1$ enhances stability and improves overall consistency. However, increasing the constraint strength to $\alpha = 0.2$ leads to a performance decline to 93.00\%. This observation suggests that excessively strong invariance regularization may over-smooth the latent manifold, thereby suppressing subtle but semantically meaningful defect cues.
    
 \textbf{Contextual Canonicalization (Variant D).} Integrating Light-Aware Linear Attention ($LA_3$) further elevates performance to 93.72\%. By aggregating global contextual information, $LA_3$ resolves ambiguities arising from localized artifacts, such as differentiating shadow boundaries from structural cracks, which purely local operators fail to disambiguate effectively.

\subsection{Active Robotic Evaluation (Addressing Q3)}
\label{subsec: Active Robotic Evaluation}

This experiment evaluates the Active Canonicalization Policy through real-world deployment on a physical robotic platform. The robotic system performs inspection tasks under adversarial illumination conditions, such as strong specular glare directly obscuring surface scratches and other subtle defects. During evaluation, the agent is permitted up to three sequential re-orientation actions. Each action is guided by epistemic uncertainty, which serves as a feedback signal for pose adjustment. This closed-loop procedure enables the robot to actively search for a geometrically favorable viewpoint that reduces condition-induced ambiguity prior to final anomaly inference.

% \textbf{Results:}
% \begin{itemize}[label=$\bullet$]
%     \item \textbf{Passive Failure:} Static methods (Dinomaly) achieve only \textbf{68.2\%} accuracy. They are fundamentally limited by the physics of reflection—no algorithm can detect a scratch that is optically invisible due to specular saturation.
%     \item \textbf{Active Success:} PiCo's active policy achieves \textbf{98.5\%} accuracy.
%     \begin{itemize}
%         \item Step 0 (Initial View): 68\% accuracy (High Uncertainty).
%         \item Step 1 (Re-orient): 85.4\% accuracy.
%         \item Step 2 (Canonical View): \textbf{98.5\%} accuracy.
%     \end{itemize}
% \end{itemize}

\begin{figure}
    \centering
    \includegraphics[width=1.0\linewidth]{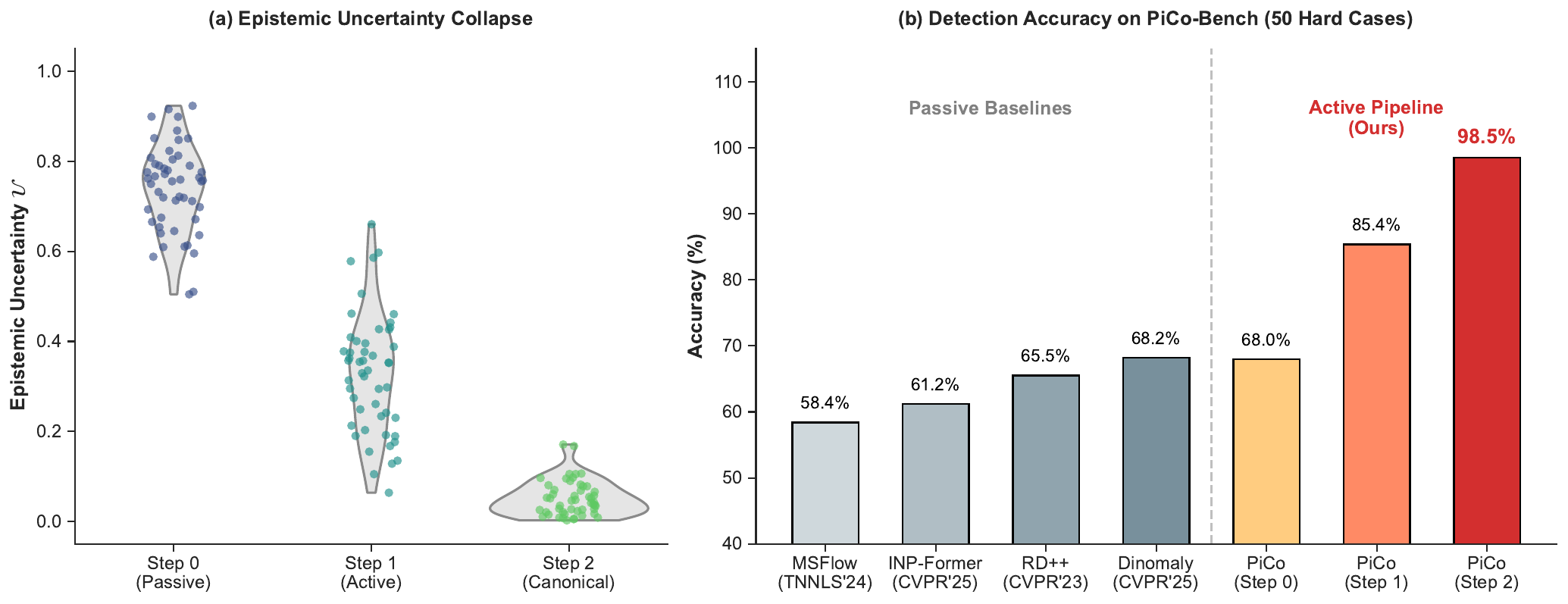}
    \caption{\textbf{Quantitative evaluation of active canonicalization on 50 hard cases.} \textbf{(a) Uncertainty collapse:} Active robotic re-orientation forces widely dispersed, high-uncertainty observations to converge into a tight, low-uncertainty manifold. \textbf{(b) Accuracy comparison:} While passive SOTA baselines fail under severe physical occlusions (peaking at 68.2\%), PiCo's active pipeline breaks this physical ceiling, achieving 98.5\% accuracy at the canonical view.}
    \label{fig:6}
\end{figure}

As illustrated in Fig. \ref{fig:6}(b), under severe physical occlusion, all passive state-of-the-art baselines suffer catastrophic performance degradation, failing to surpass the 70\% accuracy threshold.
The fundamental reason PiCo is able to break this apparent physical limit lies in the cognitive uncertainty collapse visualized in Fig. \ref{fig:6}(a). By leveraging epistemic uncertainty $\mathcal{U}$ as a real-time feedback signal to actively regulate the 6-DoF pose configuration, PiCo navigates challenging out-of-distribution observations toward a low-uncertainty manifold.
Through this closed-loop intervention, detection accuracy is dynamically elevated from an initial 68.0\% to a near-saturated 98.5\%, as shown in Fig. \ref{fig:6}(b).

This empirically validates our core argument: \textbf{Robustness in embodied perception is not just a learning problem, but a control problem.} By physically transforming intractable observation distributions into a predictable, low-entropy canonical manifold, active canonicalization fundamentally overcomes the inherent limitations of passive static vision.

%% file: 5_Con.tex
% In this work, we identified the Semantic-Nuisance Entanglement as the primary bottleneck hindering the deployment of visual anomaly detection in robotic environments. We argued that the prevailing passive perception paradigm is theoretically insufficient for dynamic physical interactions.
% To bridge this gap, we introduced PiCo, a framework founded on Active Manifold Canonicalization. PiCo unifies robotic control and representation learning into a single coherent system:
% 1. Physically, it empowers robots to actively reject geometric uncertainty via closed-loop re-orientation.
% 2. Neurally, it implements a rigorously designed Conditional Information Bottleneck, enforcing invariance via a cascaded hierarchy of Photometric, Latent, and Contextual canonicalization.
% Our extensive evaluations on $M^2AD$ and PiCo-Bench demonstrate that PiCo not only establishes a new static SOTA (93.7\% AUROC) but, more importantly, unlocks near-perfect reliability (98.5\%) in closed-loop operations. By formally treating "nuisance" not as noise to be augmented, but as a variable to be canonicalized, PiCo paves the way for the next generation of resilient, embodied industrial inspectors.

This work identifies Semantic-Nuisance Entanglement as the fundamental bottleneck limiting the deployment of visual anomaly detection in robotic environments. The prevailing passive perception paradigm is shown to be theoretically inadequate under dynamic physical interactions, where condition-induced entropy dominates semantic signals. \textit{\textbf{PiCo}} is introduced as a framework grounded in Active Manifold Canonicalization. The framework unifies robotic control and representation learning within a coherent perception--action loop. Specifically, first, at the physical level, geometric uncertainty is actively reduced through closed-loop pose reorientation, contracting the high-entropy observation distribution prior to inference. Second, at the representational level, a rigorously structured Conditional Information Bottleneck is employed to enforce invariance through a cascaded hierarchy of Photometric, Latent, and Contextual canonicalization. Extensive evaluations on $M^2\mathrm{AD}$ and \textit{\textbf{PiCo-Bench}} demonstrate that \textit{\textbf{PiCo}} not only establishes a new state-of-the-art static performance of 93.7\% AUROC, but more critically achieves 98.5\% accuracy under closed-loop active operation. By reframing nuisance variability as a controllable factor to be canonicalized rather than passively augmented, this work advances a principled pathway toward resilient embodied visual inspection in industrial settings.